\documentclass[letterpaper, 10 pt, conference]{ieeeconf}  % Comment this line out if you need a4paper

\IEEEoverridecommandlockouts                              % This command is only needed if 
\usepackage{graphics} % for pdf, bitmapped graphics files
\usepackage{epsfig} % for postscript graphics files
\usepackage{times} % assumes new font selection scheme installed
\usepackage{amsmath} % assumes amsmath package installed
\usepackage{amssymb}  % assumes amsmath package installed
\usepackage{color}
\usepackage{booktabs}
\usepackage{multirow}
\usepackage{algorithm}
\usepackage{algorithmicx}
\usepackage{algpseudocode}
\usepackage{bm}
\usepackage{xcolor}
\usepackage{booktabs}
\usepackage{makecell}

\usepackage{balance}

\usepackage{marvosym}
\usepackage{ifsym}

\usepackage[
	colorlinks,
	anchorcolor=blue,
	linkcolor=cyan,
	urlcolor=red,
	citecolor=green,
	backref=page]{hyperref}

\title{\LARGE \bf 
EMIFF: Enhanced Multi-scale Image Feature Fusion for Vehicle-Infrastructure Cooperative 3D Object Detection
}

\author{Zhe Wang\textsuperscript{1}, Siqi Fan\textsuperscript{1}, Xiaoliang Huo\textsuperscript{1,2}, Tongda Xu\textsuperscript{1}, Yan Wang\textsuperscript{1*}, Jingjing Liu\textsuperscript{1}, Yilun Chen\textsuperscript{1}, Ya-Qin Zhang\textsuperscript{1*}\\
\thanks{$^{1}$Zhe Wang, Siqi Fan, Tongda Xu, Yan Wang$^{*}$, Jingjing Liu, Yilun Chen, and Ya-Qin Zhang$^{*}$ are with the Institute for AI Industry Research (AIR), Tsinghua University, Beijing, China.
{\tt\small \{wangzhe, fansiqi, xutongda, wangyan \}@air.tsinghua.edu.cn}}
\thanks{$^{2}$Xiaoliang Huo is with the School of Software, Beihang University, Beijing, China. {\tt\small huoxiaoliangchn@buaa.edu.cn}}%
}

\begin{document}

\maketitle
\pagestyle{empty}
\thispagestyle{empty}

\renewcommand{\thefootnote}{\fnsymbol{footnote}}
\footnotetext[1]{Correspondence to Yan Wang, Ya-Qin Zhang.}
\footnotetext[2]{Code will be released at \href{https://github.com/Bosszhe/EMIFF}{https://github.com/Bosszhe/EMIFF}}

%%%%%%%%%%%%%%%%%%%%%%%%%%%%%%%%%%%%%%%%%%%%%%%%%%%%%%%%%%%%%%%%%%%%%%%%%%%%%%%%
\begin{abstract}

In autonomous driving, cooperative perception makes use of multi-view cameras from both vehicles and infrastructure, providing a global vantage point with rich semantic context of road conditions beyond a single vehicle viewpoint. Currently, two major challenges persist in vehicle-infrastructure cooperative 3D (VIC3D) object detection: $1)$ inherent pose errors when fusing multi-view images, caused by time asynchrony across cameras;  $2)$ information loss in transmission process resulted from limited communication bandwidth.
To address these issues, we propose a novel camera-based 3D detection framework for VIC3D task, \textit{Enhanced Multi-scale Image Feature Fusion} (EMIFF).
To fully exploit holistic perspectives from both vehicles and infrastructure, we propose \textit{Multi-scale Cross Attention} (MCA) and \textit{Camera-aware Channel Masking} (CCM) modules to enhance infrastructure and vehicle features at scale, spatial, and channel levels to correct the pose error introduced by camera asynchrony. We also introduce a \textit{Feature Compression} (FC) module with channel and spatial compression blocks for transmission efficiency. Experiments show that EMIFF achieves SOTA on DAIR-V2X-C datasets, significantly outperforming previous early-fusion and late-fusion methods with comparable transmission costs.

\end{abstract}

%%%%%%%%%%%%%%%%%%%%%%%%%%%%%%%%%%%%%%%%%%%%%%%%%%%%%%%%%%%%%%%%%%%%%%%%%%%%%%%%
\section{INTRODUCTION}

Subject to sensor limitations, autonomous vehicles lack a global perception capability for monitoring holistic road conditions and accurately detecting surrounding objects, which bears great safety risks~\cite{ma2022bevsurvey,han2023collaborative}. Vehicle-to-everything (V2X)~\cite{xu2022opv2v,mehr2019disconet} aims to build a communication system between vehicles and other devices in a complex traffic environment. Vehicle and infrastructure cooperation can significantly expand the perception range and improve perception capability~\cite{yu2022dairv2x}. Cameras from both two sides provide a global vantage point with a rich semantic context of road conditions beyond a single-vehicle viewpoint~\cite{xu2022cobevt}. Vehicle-infrastructure cooperative 3D object detection (VIC3D) from cameras is a significant task for autonomous driving.

\begin{figure}[t]
	\centering  
	\includegraphics[width=0.85\linewidth]{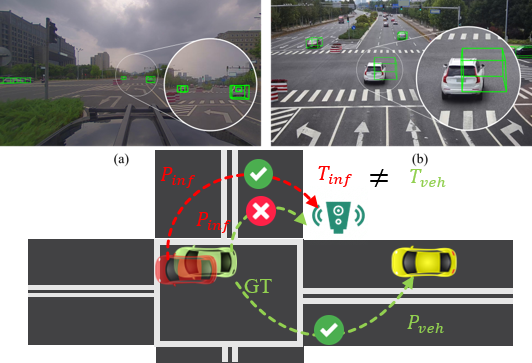} 
	\caption{Labels (3D bounding boxes) projected from 3D space to vehicle (a)  and infrastructure (b) image planes using calibration parameters $P_{inf/veh}$ often suffer from misalignment between the ground truth and the projection position in 2D images (as illustrated by the misaligned green bounding boxes).  The reason for this misalignment is that the camera's capture time $T_{inf/veh}$ are different and the moving object captured from the vehicle camera (in green) and infrastructure camera (in red) will appear at different locations.}  
	\label{fig:calib_noise}   
\end{figure}

Compared with vanilla single-vehicle 3D object detection, VIC3D tasks face more unique challenges. One challenge is inherent pose error when fusing multi-view images from vehicles and those from infrastructure~\cite{CorrectPE}, caused by time asynchrony across agents~\cite{yu2023ffnet}. As shown in Figure~\ref{fig:calib_noise}, this pose errors can result in inaccurate relative positions between objects and annotations. Another challenge is limited communication bandwidth between agents resulting in information loss between transmissions~\cite{xu2022opv2v}. The raw sensor data possesses ample information required for fusion; however, it necessitates greater bandwidth, thus necessitating fusion methods to prioritize the balance between performance and transmission cost. Therefore, fusion methods to tackle such cross-agent perception challenges are the key to VIC3D.

Many fusion works on V2X are proposed based on simulated datasets, such as OPV2V~\cite{xu2022opv2v}, V2X-Sim~\cite{li2022v2xsim} and V2XSet~\cite{xu2022v2xvit}, which neglect above challenges and have a sim-to-real gap.  Most existing research only focused on LiDAR-based methods due to the fusion convenience and the performance advantage, such as \textit{early fusion} (EF) of raw signals~\cite{yu2022dairv2x,hu2022where2comm,chen2022co3}, \textit{intermediate fusion} (IF) of features~\cite{mehr2019disconet,xu2022opv2v,wang2020v2vnet,fan2023quest}, and \textit{late fusion} (LF) of prediction outputs~\cite{yu2022dairv2x,chen2022model-agnostic}. But due to the projection gap between 2D image plane and 3D space, image fusion can not be as direct as point clouds. In real scenarios, DAIR-V2X~\cite{yu2022dairv2x} adopts an LF method by combining prediction outputs from each camera, which is sensitive to calibration so that even when prediction from the infrastructure side is perfect, the vehicle will receive biased 3D detection.

In this paper, we propose a novel framework for VIC3D task, \textit{Enhanced Multi-scale Image Feature Fusion} (EMIFF). We choose intermediate fusion since it doesn't highly rely on accurate calibration parameters. For feature-level fusion, high-dimensional features extracted from raw data can be compressed, transmitted, and dynamically enhanced~\cite{yu2023ffnet}, which can be used to alleviate the negative effect of pose errors. We design modules to compress transmitted features to reduce transmission cost and enable feature enhancement in scale level, spatial level, and channel level.

Specifically, Feature Compression (FC) module compresses 2D features transmitted from the infrastructure to vehicle. Since the receptive field is larger in smaller-scale features, which theoretically has higher tolerance to slight location errors, Multi-scale Cross Attention (MCA) module aims to achieve attentive scale-wise feature selection between featurese. MCA also corrects features at the spatial level with attentive offset to overcome pixel-wise shift caused by pose errors. To correct location errors born from multiple cameras, features are further enhanced by a  Camera-aware Channel Masking (CCM) module via a learned channel-wise mask following the guidance of camera parameters. Then, the enhanced features are transformed into voxel features leveraging calibration parameters. Finally splatted into BEV space, the fused feature is fed into detection heads for object detection. Experiments demonstrate the effectiveness of each EMIFF module in reducing pose errors and achieving better prediction accuracy than existing EF and LF methods. Our contributions can be summarized as follows:

\begin{itemize}
    \item We propose EMIFF, a novel framework for camera-based VIC3D object detection, using an intermediate fusion method to tackle cross-agent perception challenges.
    \item We design MCA and CCM modules to dynamically enhance image features for better detection performance, with an additional FC module to reduce transmission costs in VIC3D system.
    \item We achieve state-of-the-art results on DAIR-V2X-C dataset, the latest VIC3D benchmark with real data, where EMIFF outperforms existing LF and EF methods with comparable transmission costs. 
\end{itemize}

\section{Related Work}
\subsection{V2X Cooperative Perception}

%%%%%%%%%%%%%%%%%%%%%%%%%%%%%%%%%%%%%%%%%%%%%%%%%
% Many fusion works on V2X field are proposed based on simulated datasets, such as OPV2V~\cite{xu2022opv2v}, V2X-Sim~\cite{li2022v2xsim} and V2XSet~\cite{xu2022v2xvit}, which ignores real challenges and have a huge sim-to-real gap.  Most existing research on VIC3D has focused on LiDAR-based methods due to the fusion convenience and the performance advantage, such as \textit{early fusion} (EF) of raw signals~\cite{yu2022dairv2x,hu2022where2comm,chen2022co3}, \textit{intermediate fusion} (IF) of features~\cite{mehr2019disconet,xu2022opv2v,wang2020v2vnet}, and \textit{late fusion} (LF) of prediction outputs~\cite{yu2022dairv2x,chen2022model-agnostic}. As LiDAR is highly expensive and difficult to deploy in each vehicle in practical applications, an alternative solution is Vehicle-to-Infrastructure (V2I), in which case standard cameras are installed in shared traffic environment providing a holistic view of road conditions. But due to the gap between 2D image plane and 3D space, image fusion can not be as direct as point clouds.

Current research on V2X cooperative perception mainly focuses on simulated datasets, such as OPV2V~\cite{xu2022opv2v}, V2X-Sim~\cite{li2022v2xsim} and V2XSet~\cite{xu2022v2xvit}. Existing intermediate-fusion methods focused on simulated point clouds, such as V2VNet~\cite{wang2020v2vnet} which transmitted compressed features to nearby vehicles and generated joint perception/prediction. DiscoNet~\cite{mehr2019disconet} introduced graphs into feature fusion and proposed edge weights to highlight different informative regions during feature propagation. Recent Where2comm~\cite{hu2022where2comm} considered the spatial confidence of features and selected features with high confidence and complementary to others, which effectively saves transmission costs. FFNET~\cite{yu2023ffnet} introduces the concept of feature flow to tackle the issue of temporal asynchrony in real-world scenarios, albeit its applicability is limited to point cloud data.
Different from point clouds, images from vehicle and infrastructure have a huge view gap, thus features need to be transformed into unified space for fusion. One direct way for fusing multi-view images is late fusion, such as DAIR-V2X~\cite{yu2022dairv2x}, which proposed a result-level fusion model for cameras with separate detectors~\cite{rukhovich2022imvoxelnet}. Few approaches have focused on IF methods for cameras, especially in real scenarios. 
%, while cameras can provide extra and complementary visual information compared to point clouds, making it essential for further V2X studies.

\begin{figure*}[t]
	\centering  
	\includegraphics[width=0.9\linewidth]{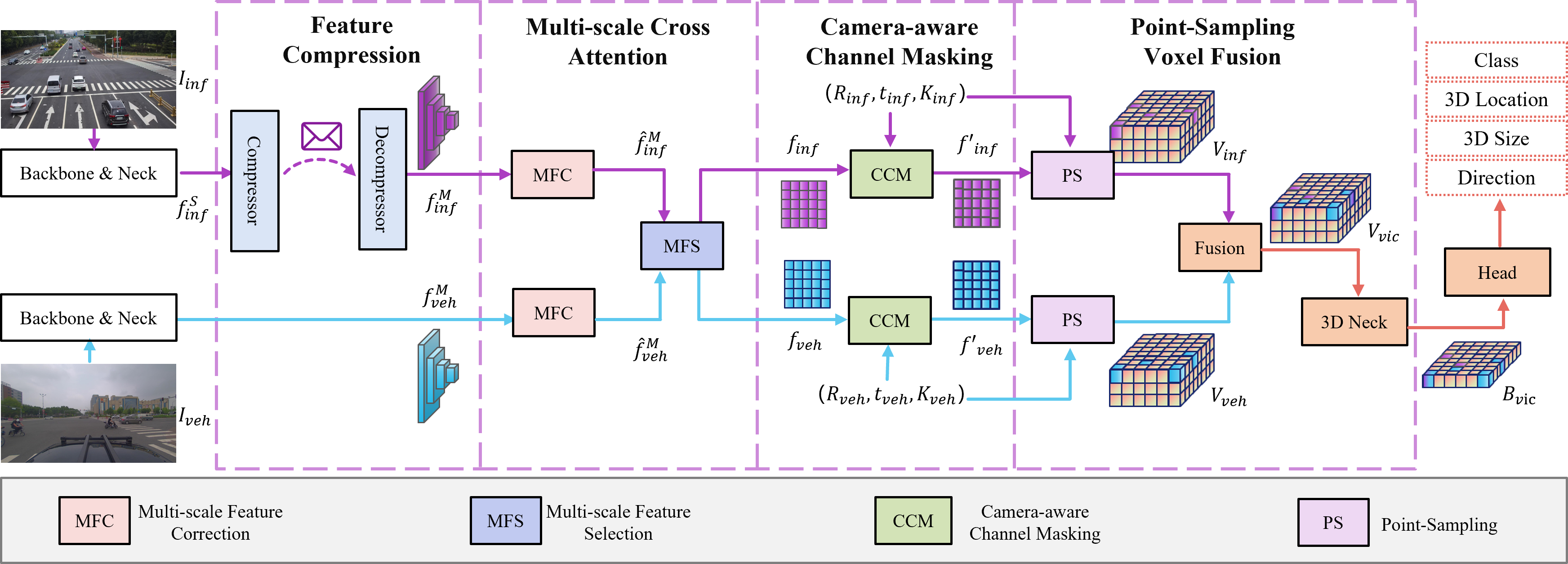} 
	\caption{The general framework of EMIFF. Separate image backbone and neck extract multi-scale image feature from vehicle and infrastructure images. FC module compresses source infrastructure feature $f^{S}_{inf}$ and decompresses it to multi-scale ones $f^{M}_{inf}$. MCA module consisting of MFC and MFS blocks enhances multi-scale features $f^{M}_{veh/inf}$ by seeking the correlation between the two sides, and CCM takes camera parameters $(R,t,K)$ as input to reweight features $f_{veh/inf}$ with channel relationship. Finally, Point-Sampling Voxel Fusion projects image features $f^{\prime}_{veh/inf}$ into 3D space to generate a unified voxel feature $V_{vic}$, which can be applied to 3D neck and head in turn for detection prediction. 
 }  
	\label{fig:framework}   
\end{figure*}

\subsection{Camera-based Feature Fusion}

%Muti-view camera fusion methods can be summarized into three main categories: direct prediction, lift-based, and projection-based methods.

    \textbf{Direct Prediction} methods extract image features with object query~\cite{wang2022detr3d,chen2022futr3d,liu2022petr,liu2022petrv2} or directly on front-view image~\cite{wang2021fcos3d}. DETR3D~\cite{wang2022detr3d} used a sparse set of 3D object queries to sample 2D multi-view image features and predicted 3D bounding boxes with set-to-set loss. PETR~\cite{liu2022petr,liu2022petrv2} transformed image features into 3D position-aware representation by encoding 3D coordinates into position embedding. FCOS3D~\cite{wang2021fcos3d} transformed 3D labels to front-view images and directly predicted 3D information by extending FCOS~\cite{tian2019fcos} to 3D detection.

\textbf{Lift-based} methods project features from image plane to BEV  (bird's eye view) plane through depth estimation. Most methods~\cite{huang2021bevdet,huang2022bevdet4d,xie2022m2bev,zhang2022beverse,reading2021caddn} applied 2D-to-3D transformation following LSS~\cite{philion2020lss}, which predicted a depth distribution for each pixel and lifted image features into frustum features with camera parameters, then splatted all frustums into a rasterized BEV feature. BEVDepth~\cite{li2022bevdepth} claimed the quality of intermediate depth estimation is the key to improving multi-view 3D object detection and added explicit depth supervision with groundtruth depth generated from point clouds. PON~\cite{roddick2020pon} learned the transformation leveraging geometry relationship between image locations and BEV locations in the horizontal direction. 

\textbf{Projection-based} methods generate dense voxel or BEV representation from image features through 3D-to-2D projection~\cite{ma2022bevsurvey}. ImVoxelNet~\cite{rukhovich2022imvoxelnet} aggregated the projected features from several images via a simple element-wise averaging, where spatial information might not be exploited sufficiently.
Transformer-based methods~\cite{li2022bevformer,peng2022bevsegformer} mapped perspective view to BEV with designed BEV queries and leveraged cross- and self-attention to aggregate spatial and temporal information into BEV queries. Since global attention needs huge memory with high time cost, deformable attention was adopted in BEVFormer~\cite{li2022bevformer}.

\section{Method}

EMIFF aims to fuse vehicle and infrastructure features by utilizing V2X communication. It includes four main modules: Feature Compression (FC), Multi-scale Cross Attention (MCA), Camera-aware Channel Masking (CCM), and Point-Sampling Voxel Fusion, as illustrated in Figure~\ref{fig:framework}.

\begin{figure}[ht]
	\centering  
        \includegraphics[width=0.9\linewidth]{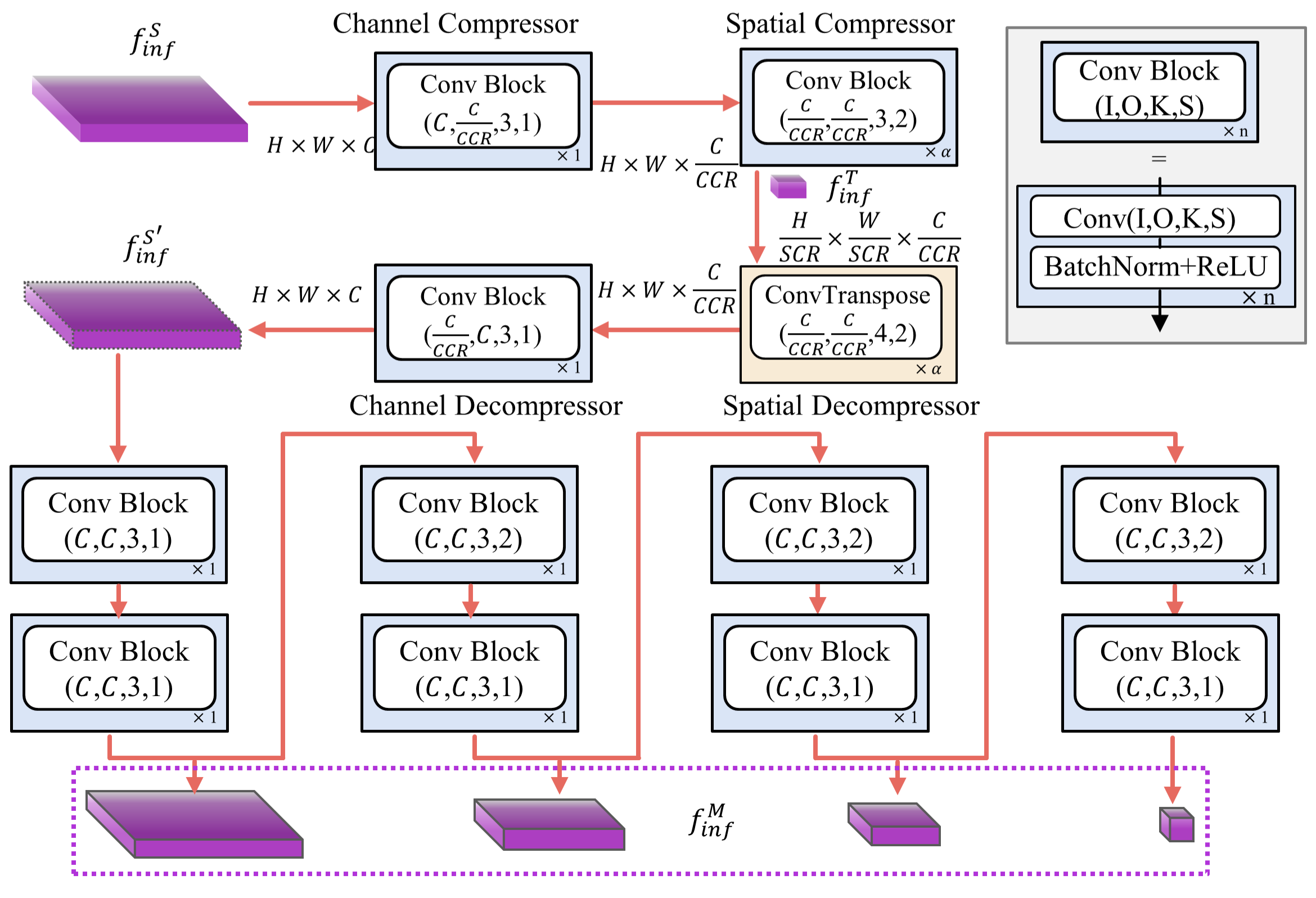} 
	\caption{Illustration of FC module. Feature $f_{inf}^{S}$ is compressed into $f_{inf}^{T}$ through the channel and spatial compressors, which is transmitted to vehicle and is decoded into $f_{inf}^{S\prime}$ through the channel and spatial decompressors. Finally, multi-scale infrastructure features $f_{inf}^{M}$ can be recovered from $f_{inf}^{S\prime}$ with several Conv Blocks with stride 2. 
 }
	\label{fig:FC}   
\end{figure}

\subsection{Feature Compression}

The images from vehicle and infrastructure are denoted as $I_{veh}$ and $I_{inf}$, respectively, and the shape of both images are $ \left[ H \times W \times 3 \right]$. Since infrastructure cameras are typically installed at a higher elevation than vehicles, resulting in a huge view gap between images captured from vehicles and infrastructure, we use separate pre-trained backbones and necks on the vehicle and infrastructure respectively to extract multi-scale image features. The output multi-scale features can be denoted as $f^{M}_{s}, s = veh/inf$.

EMIFF transmits image features and camera parameters instead of voxel feature after projection because voxel feature is too large to be transmitted efficiently. The Feature Compression (FC) module (shown in Figure~\ref{fig:FC}) compresses the largest infrastructure feature $f^{0}_{inf}$ (noted as $f^{S}_{inf}$) to  $f_{inf}^{T}$, transmits  $f_{inf}^{T}$ to vehicle and regenerate multi-scale features$f^{M}_{inf}$ through decompression.

% The compression and decompression process in FC module is an Encoder-Decoder with four components: Channel Compressor (CC), Spatial Compressor (SC), Spatial Decompressor (SD), and Channel Decompressor (CD). CC and CD are composed of several convolutional layers. SC comprises several Conv Blocks with stride 2 so that feature scales are reduced to half after each one. SD only replaces convolution with transposed convolution. The Compression Rate (CR) is determined by Channel Compression Rate (CCR) and Spatial Compression Rate (SCR). The number of SC's layers is calculated by $\alpha = \log_{4}{\text{SCR}}$. 

\subsection{Multi-scale Cross Attention}

\begin{figure}[ht]
	\centering  
	\includegraphics[width=0.9\linewidth]{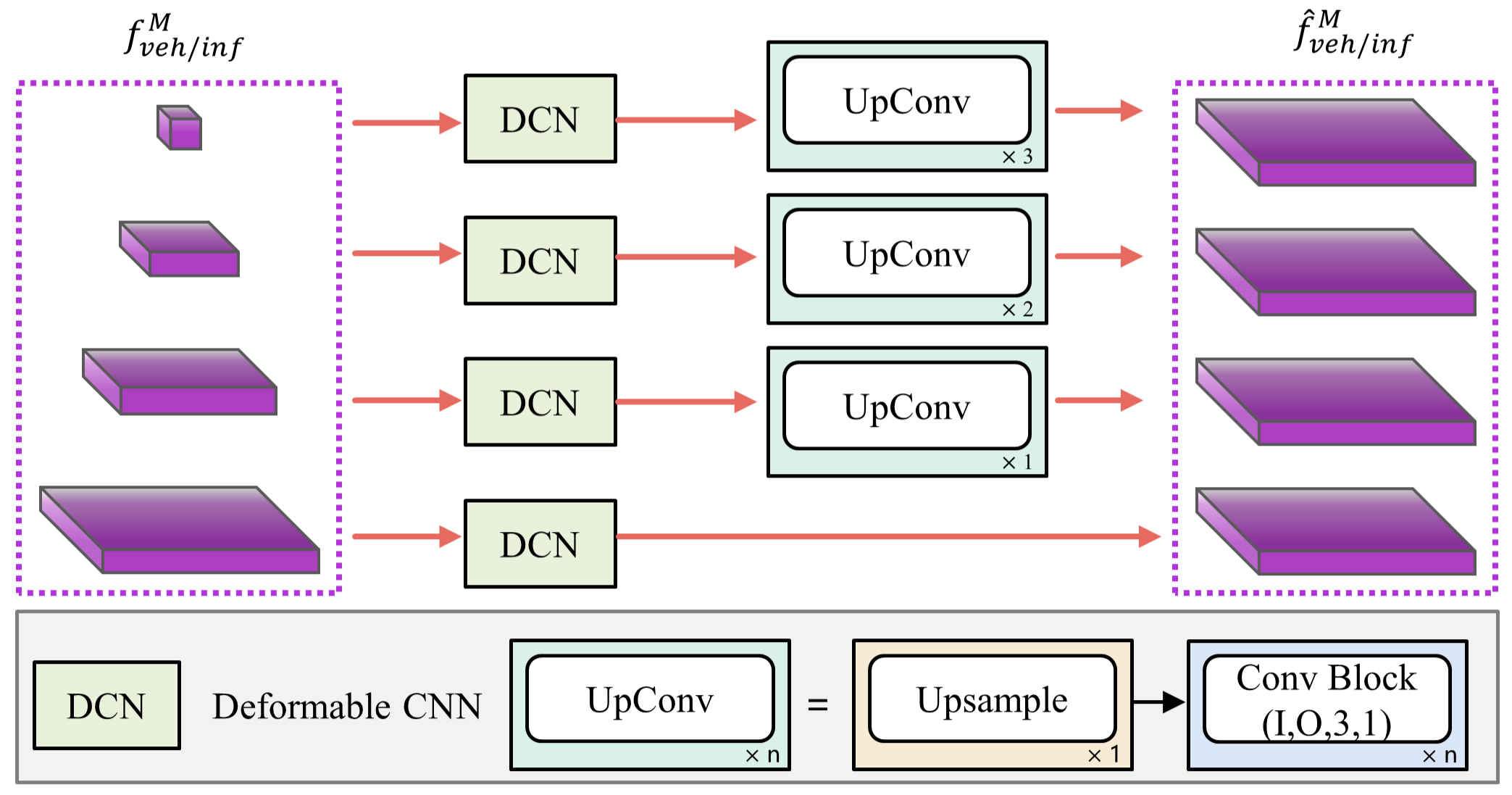} 
	\caption{Details of MFC. Every pixel-wise feature is integrated with the spatial information of surrounding pixels via DCN, and multi-scale features are scaled to the same size through UpConv blocks.}  
	\label{fig:MFC}   
\end{figure}

\begin{figure}[ht]
	\centering  
	\includegraphics[width=0.9\linewidth]{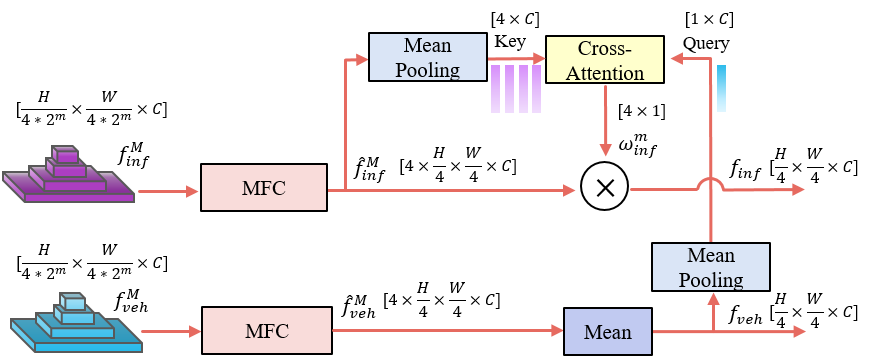} 
	\caption{Schema of MCA module.  In the lower branch, vehicle feature $f_{veh}$ is generated from $f^{M}_{veh}$ through MFC Block and Mean. In the upper branch, $f^{M}_{inf}$ is refined into `key' through MFC Block and MeanPooling, and queries are generated from $f_{veh}$ through MeanPooling. The output weights $\omega_{inf}^{m}$ of cross-attention are applied to $\hat{f}^{M}_{inf}$ with inner product to form infrastructure feature$f_{inf}$.} 
	\label{fig:MCA}   
\end{figure}

MCA module contains Multi-scale Feature Correction (MFC) and  Multi-scale Feature Selection (MFS) blocks, as shown in Figure~\ref{fig:MFC} and Figure~\ref{fig:MCA}. MFC block is designed to select and integrate spatial-wise features with attentive offset, and thus pixel-wise shift caused by pose errors can be overcome to some extent. MFS block applies cross-attention between them to achieve attentive scale-wise feature selection.
 
The MFC module is first applied to multi-scale features. Since pose errors can cause a displacement between the projected and ground-truth positions on 2D plane, we apply DCN (deformable convolutional networks) \cite{Dai_2017_dcn} for each scale feature to allow every pixel to get spatial information surrounding it. Then, features at different scales are upsampled to the same size through UpConv blocks.

 MFS applies MeanPooling operation to obtain the representation of different scales of infrastructure features, while vehicle features at different scales are first fused by mean operation and then refined by MeanPooling. To find the correlation between vehicle features and infrastructure features at different scales, cross attention is applied to infrastructure representations as Key and vehicle representation as Query, which generates attention weights $\omega^{m}_{inf}$ for each scale $m$. We calculate inter-product between features $\hat{f}^{M}_{inf}$ and weights $\omega^{m}_{inf}$. The final outputs of MCA are augmented infrastructure image feature $f_{inf}$ and vehicle image feature $f_{veh}$.

% \begin{figure}[htbp]
% \subfigure[Details of MS Block]{
% \begin{minipage}[t]{0.48\textwidth}
% 	\centering  
% 	\includegraphics[width=\linewidth]{Figures/MSB.png} 
%  	% \caption{Details of MS Block.}  
%   	\label{fig:MSB}  
%  \end{minipage}
%  }
%  \subfigure[Schema of MCA]{
%  \begin{minipage}[t]{0.48\textwidth}
%         \includegraphics[width=\linewidth]{Figures/MCA.png} 
% 	% \caption{Schema of MCA module.} 
% 	\label{fig:MCA}   
%  \end{minipage}
%  }

%  \caption{Illustration of MS Block and MCA modules. Every pixel-wise feature is integrated with the spatial information of surrounding pixels via DCN, and multi-scale features are scaled to the same size through UpConv blocks. In the lower branch of MCA, vehicle feature $f_{veh}$ is generated from $f^{M}_{veh}$ through MS Block and Mean. In the upper branch, $f^{M}_{inf}$ is refined into `key' through MS Block and MeanPooling, and queries are generated from $f_{veh}$ through MeanPooling. The output weights $\omega_{inf}^{m}$ of cross-attention are applied to $\hat{f}^{M}_{inf}$ with inner product to form infrastructure feature$f_{inf}$.}
% \end{figure}

\subsection{Camera-aware Channel Masking}

Considering that objects closer to the camera are easier to detect and it’s common for the same object to be closer to an infrastructure camera but far away from a vehicle. Since different channels represent object information at different distances, which is strongly correlated with camera parameters, it is intuitive to take camera parameters as priors to augment image features.

Inspired by the decoupled nature of SENet~\cite{hu2018SE} and  LSS~\cite{philion2020lss},  CCM will learn a channel-wise mask to weigh the importance between the channels. First, camera intrinsic and extrinsic are stretched into one dimension and concatenated together. Then, they are scaled up to the feature’s dimension $C$ using MLP to generate a channel mask $M_{veh/inf}$. Finally, $M_{veh/inf}$ is used to re-weight the image features $f_{veh/inf}$ in channel-wise and obtain results $f^{\prime}_{veh/inf}$. The overall CCM module can be written as:

\begin{equation}
    \begin{split}
        f^{\prime}_{s}  &= M_{s} \odot f_{s} , s= veh, inf \\
        m_{s} &=  \text{MLP} \left(\xi\left(R_s\right) \oplus \xi\left(t_s\right) \oplus \xi\left(K_s\right)\right)
    \end{split}
\end{equation}

$\xi$ denotes the flat operation and $\oplus$ means concatenation. The input of MLP is the combination of camera rotation matrix $R_s\in \mathbb{R}^{3\times3}$, translation $t_{s}$ and camera intrinsics $K_{s}$. $M_{s}$ can be obtained from MLP's output $m_{s}$ through several Fully Connected (FC) layers and Activation (Relu, Sigmoid) layers. 

\begin{figure}[h]
	\centering  
	\includegraphics[width=0.8\linewidth]{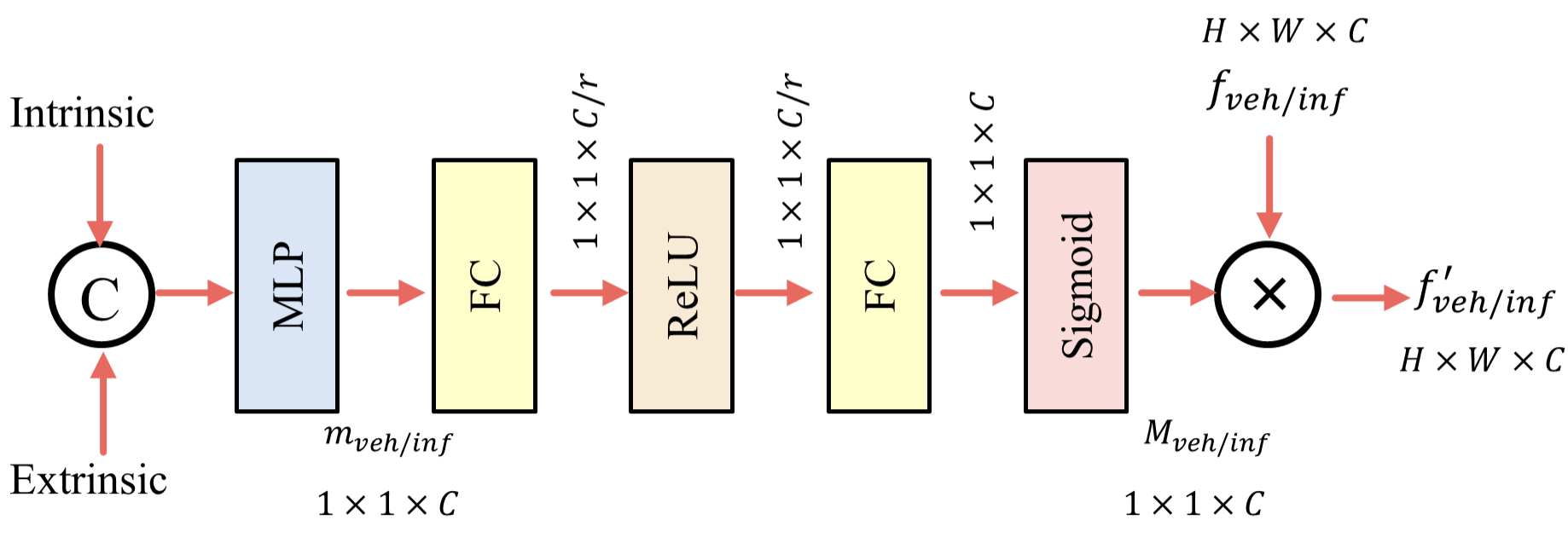} 
	\caption{The schema of CCM module. 
 % Camera intrinsic and extrinsic are encoded into the channel mask, then image feature is integrated with it through inner-product operation.
 }  
	\label{fig:CCM}   
\end{figure}

\subsection{Point-Sampling Voxel Fusion}

The augmented vehicle feature $f^{\prime}_{veh}$ and infrastructure feature $f^{\prime}_{inf}$ are projected into 3D space for fusion and generate two voxel features, denoted as $V_{veh}$ and $V_{inf}$, respectively. The details of projection principle are following ImVoxelNet ~\cite{rukhovich2022imvoxelnet}.

We obtain the final voxel feature $V_{vic} \in N_x\times N_y\times N_z \times C_1$ by averaging sampled voxel features $V_{veh}$ and $V_{inf}$. Then,  the same 3D neck as~\cite{rukhovich2022imvoxelnet}, which is composed of 3D CNN and downsampling layers, transforms voxel feature $V_{vic}$ into BEV feature $B_{vic} \in N_X \times N_y \times C_2$. BEV feature can be used as input of common 2D detection heads to predict 3D detection results. The loss of detection heads is similar to SECOND~\cite{yan2018second}, which consists of smooth L1 Loss for bounding box $L_{\text{bbox}}$, focal loss for classification $L_{\text{cls}}$, and cross-entropy loss for direction $L_{\text{dir}}$.

\section{Expermients}

\begin{figure*}[t]
	\centering  
	\includegraphics[width=0.85\linewidth]{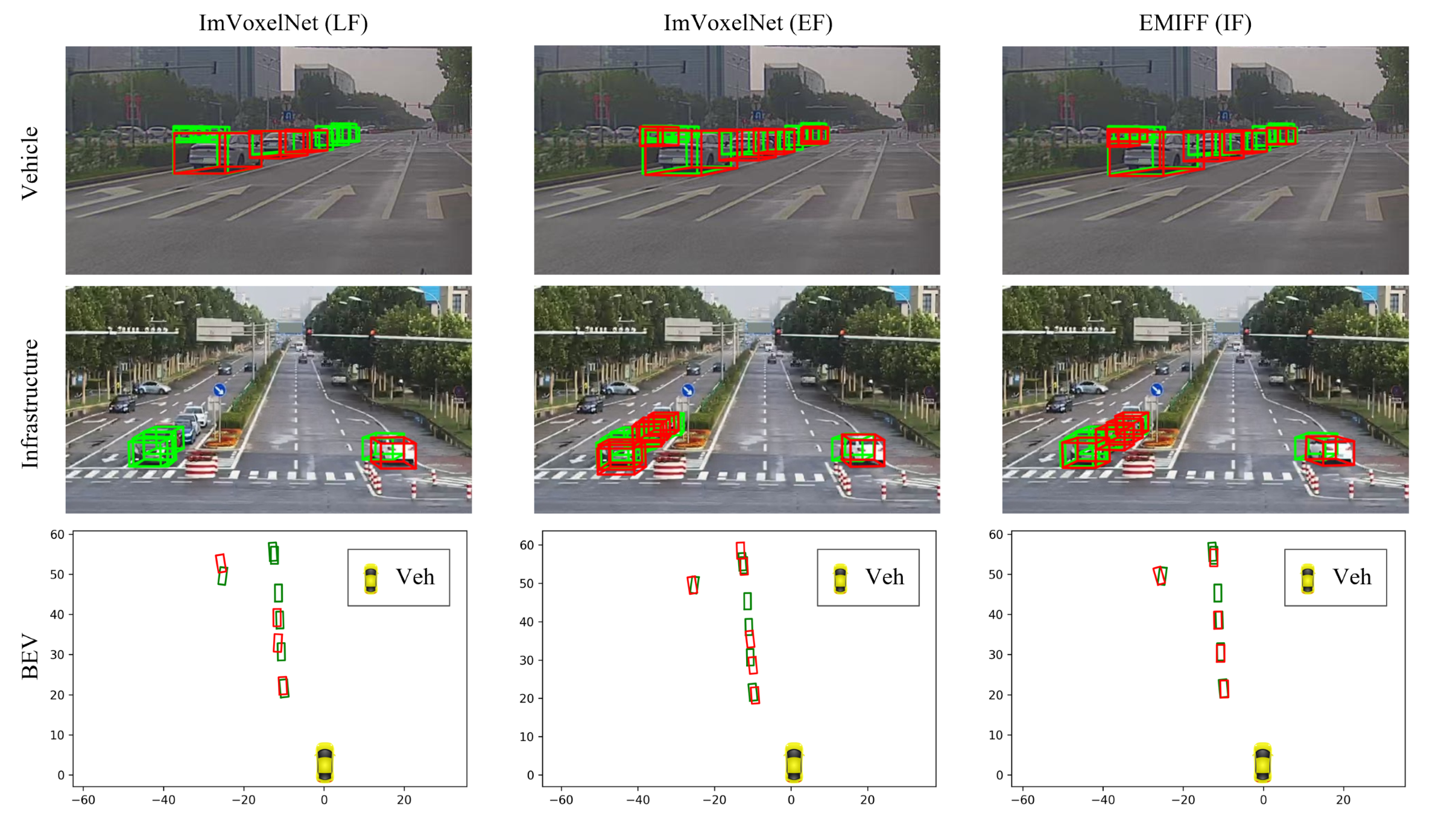} 
	\caption{Visualization results of ImVoxelNet (LF) (left column), ImVoxelNet\_M (EF) (middle column), and EMIFF (IF) (right column). Bounding boxes in BEV (bottom row) are projected to vehicle and infrastructure image planes (top two rows). Groundtruth are in green and predictions in red. 
 From BEV, it is clear that red and green bounding boxes from EMIFF are better aligned than LF and EF methods. This shows that ImVoxelNet (LF) and  ImVoxelNet\_M (EF) have detected more false positive objects and fewer true positive objects than EMIFF (IF).}
	\label{fig:vis_results}   
\end{figure*}

\subsection{Implementation Details}

\textbf{Datasets.}
We conduct our experiments on a vehicle-infrastructure-cooperation dataset DAIR-V2X~\cite{yu2022dairv2x}, in which all frames are captured from real scenarios. We utilize the VIC-Sync portion of DAIR-V2X-C dataset for training and evaluation, which is composed of 9,311 pairs of infrastructure and vehicle frames captured at the same time. Annotations of each pair frame are in world coordinate and need to be translated into vehicle coordinate system for training and evaluation. 

\textbf{Evaluation Metrics.} Evaluation metrics are Average Precision (AP)~\cite{Geiger2013KITTI} and Average Byte (AB) to measure detection performance and transmission cost, the same as~\cite{yu2022dairv2x}. The AP metrics are based on the detection range surrounding the vehicle, including Overall (0-100m), 0-30m, 30-50m, and 50-100m. All metrics are calculated with $\text{IoU}=0.5$ and can be divided into 2 parts:  $AP_{\text{3D}}$ and $AP_{\text{BEV}}$. AB means the average size of transmitted data. It is the feature map $f_{inf}^{\text{T}}$ in our method.

\textbf{Training.} We use ResNet-50~\cite{he2016resnet} as backbone and FPN~\cite{lin2017fpn} as 2D neck to extract image features. The channel number $C$ of the neck's output is 64. We set the channel of 3D voxel feature $C_1$ to 64 and the channel of BEV feature $C_2$ to 256 following~\cite{yan2018second,lang2019pointpillars}.

\subsection{Object Detection Results}
\label{detection_results}

We compare the performance of baseline Late Fusion (LF) methods with ImVoxelNet and our proposed single-side model \textit{EMIFF\_Veh/Inf} on DAIR-V2X-C dataset. We also implement several multi-view camera-based methods that have been applied to nuScenes dataset~\cite{nuscenes2019,Geiger2013KITTI} (eg., BEVFormer~\cite{li2022bevformer}, BEVDepth~\cite{li2022bevdepth}) for VIC3D task . The evaluation results on VIC-Sync portion of DAIR-V2X-C dataset are shown in Table~\ref{tab:EMIFF_SOTA} and Figure~\ref{fig:vis_results}. From the table, Intermediate Fusion (IF) method EMIFF has achieved state-of-the-art performance on the multi-view camera fusion benchmark, compared with other methods of Late Fusion (LF) and Early Fusion (EF). EMIFF obtains 15.61 $AP_{\text{3D}}$ and 21.44 $AP_{\text{BEV}}$ in overall setting.

\textit{EMIFF\_Veh} and \textit{EMIFF\_Inf} remove the MCA module but preserve CCM and FC modules so that models can be applied to the vehicle side and infrastructure side respectively without interaction between them, and predictions can be used for Late Fusion. EMIFF achieves higher $AP_{\text{3D}}$ and $AP_{\text{BEV}}$ compared with ImVoxelNet~\cite{rukhovich2022imvoxelnet} under the setting of Only-Veh, Only-Inf, and LF. This indicates that EMIFF's  single-side model has a stronger feature extraction ability.

What is interesting is that Only-Inf methods achieve the best scores in 50-100m $AP_{\text{3D}}$ and $AP_{\text{BEV}}$ and this phenomenon can also be seen in~\cite{yu2022dairv2x}. As mentioned before, these metrics are related to detecting objects far from the ego vehicle. We count 16,934 objects within the distance range of 50-100m from vehicle, which are used to calculate  the metric of 50-100m $AP_{\text{3D}}$. Among these objects, almost three-quarters (12,651) objects are closer to infrastructure camera, which are easier to be detected by Only-Inf models. %brings more benefits for leveraging the infrastructure viewpoint.

\begin{table*}[t]
  % \footnotesize
  \centering
  % \resizebox{\textwidth}{!}{
    \begin{tabular}{ccccccccccc}
    \hline
    \multirow{2}{*}{\textbf{Fusion}} & \multirow{2}{*}{\textbf{Model}} & \multicolumn{4}{c}{\bm{$AP_{\textbf{3D (IoU=0.5)}}$}} & \multicolumn{4}{c}{\bm{$AP_{\textbf{BEV (IoU=0.5)}}$}} & 
    \multirow{2}{*}{\makecell[c]{AB\\(Byte)}} \\ \cline{3-10}
                                &               & Overall   & 0-30m    & 30-50m    & 50-100m   & Overall   & 0-30m     & 30-50m    & 50-100m   & \\ \hline
    \multirow{2}{*}{Only-Veh}   & ImVoxelNet~\cite{rukhovich2022imvoxelnet}    & 7.29      & 16.98     & 2.35      & 0.13      & 8.85      & 19.89     & 3.44      & 0.28      & 
                                \multirow{2}{*}{\textbackslash{}}       \\
                                & EMIFF\_Veh  & 8.65     & 19.11     & 4.33      & 0.20      & 10.46     & 22.42     & 5.57      & 0.42      & \\ \hline
    \multirow{2}{*}{Only-Inf}   & ImVoxelNet~\cite{rukhovich2022imvoxelnet}    & 8.66      & 13.05     & 5.79      & \underline{5.50}      & 14.41     & 17.98     & 10.34     & \underline{11.19}     &    
                                \multirow{2}{*}{\textbackslash{}}        \\ 
                                & EMIFF\_Inf  & 9.76     & 13.59     & 6.90      & \textbf{6.63} & 14.81 & 18.78     & \underline{11.50}     & \textbf{11.43} &  \\ \hline
    \multirow{2}{*}{LF} & ImVoxelNet~\cite{yu2022dairv2x} & 11.08  & 22.27     & 4.40      & 2.33      & 14.76     & 27.02     & 7.13      & 4.73      & 0.28K \\
                                & EMIFF\_Veh/Inf & 11.99 & \underline{24.79}     & 6.08      & 2.30      & 15.79     & 30.39     & 8.50      & 4.84      & 0.28K \\ \hline
    \multirow{3}{*}{EF}  & BEVDepth~\cite{li2022bevdepth} & 7.36    & 16.23     & 1.79      & 0.18      & 13.17     & 26.42     & 5.00      & 4.82      &  \multirow{3}{*}{550.84K} \\ 
                                & BEVFormer\_S~\cite{li2022bevformer}     & 8.80      & 18.07     & 3.71      & 1.76      & 13.45     & 24.76     & 6.46      & 4.63      &       \\
                                & ImVoxelNet~\cite{rukhovich2022imvoxelnet}    & \underline{12.72}     & 23.63     & \underline{7.38}      & 3.11      & \underline{18.17}     & \underline{30.54}     & 11.39     & 7.00      &        \\ \hline
    \multirow{1}{*}{IF}   
    & EMIFF   & \textbf{15.61} & \textbf{29.12} & \textbf{9.07} & 4.01    & \textbf{21.44} & \textbf{36.24}   & \textbf{13.51}     & 8.28      & 32.64K  \\ \hline
\end{tabular}
% }
    \caption{Quantitative evaluation on DAIR-V2X-C. Best values are marked by bold, and the second best is underlined. All scores in $\%$. 
    }
    \label{tab:EMIFF_SOTA}
\end{table*}

We also compared EMIFF with some representative cooperative perception models on DAIR-V2X-C datasets and experimental results are reported in Table~\ref{TAB:VIMI_V2X}. The performance advantage of EMIFF is significant and it outperforms DiscoNet~\cite{mehr2019disconet} by 23.69\%. ($*$ means results are from paper~\cite{hu2023coca3d}).

\begin{table}[ht]
    % \footnotesize
    \centering
    % \resizebox{\linewidth}{!}{
    \begin{tabular}{ccc} 
    \hline
    Fusion Modality& Model & $AP_{3D}$ \\
    \hline
    Bounding Box & LateFusion~\cite{yu2022dairv2x} \textcolor{blue}{(CVPR'22)} & 11.08 \\ \hline
    \multirow{6}{*}{BEV} & CoBEVT~\cite{xu2022cobevt}  \textcolor{blue}{(CoRL'22)} & 4.80 \\
    & V2VNet*~\cite{wang2020v2vnet}  \textcolor{blue}{(ECCV'20)} & 8.47 \\
    & When2com*~\cite{Liu2020when2com}  \textcolor{blue}{(CVPR'20)} & 9.84 \\
    & Where2comm*~\cite{hu2022where2comm}  \textcolor{blue}{(NeurIPS'22)} & 10.25 \\
    & V2X-ViT*~\cite{xu2022v2xvit}  \textcolor{blue}{(ECCV'22)}& 10.75 \\
    & DiscoNet*~\cite{mehr2019disconet} \textcolor{blue}{(NeurIPS'21)} & 12.62 \\ \hline
    Voxel & EMIFF & 15.61 \\
    \hline
    \end{tabular}
    \caption{Comparison results of representative cooperative methods on DAIR-V2X-C. All scores in $\%$.}
    \label{TAB:VIMI_V2X}
\end{table}

\subsection{Ablation Study}

We remove MCA, CCM, and FC modules in EMIFF and regard it as baseline in the ablation study. We also conduct experiments to investigate when to fuse information from vehicle and infrastructure.

\begin{table}[ht]
  \footnotesize
  \centering
    \begin{tabular}{ccccc}
    \hline
    \textbf{MCA} & \textbf{CCM} & \textbf{FC} & \bm{$AP_{\textbf{3D}}$} & \bm{$AP_{\textbf{BEV}}$} \\ \hline 
            &               &               & 13.60 & 20.05   \\ 
                & \checkmark    &               & 13.98 & 20.23   \\ 
     \checkmark  &               &               & 14.65 & 20.64   \\
     \checkmark  &  \checkmark    &               & \underline{15.27} & \underline{21.03}  \\
     \checkmark  &  \checkmark    &  \checkmark    & \textbf{15.61} & \textbf{21.44}   \\ \hline
     
    \end{tabular}
    \caption{Ablation study on EMIFF.}
    \label{TAB:AB}
\end{table}

\textbf{Effect of Each Component.} The ablation results on MCA, CCM, and FC modules are summarized in Table~\ref{TAB:AB}. The 1st row model can be denoted as \textit{EMIFF\_B}, which removes MCA, FC, and CCM modules and only keeps the fusion methodology at feature level. Comparing the 2nd and 3rd rows with the 1st row, both MCA and CCM can improve performance over baseline, and MCA has increased $AP_{\text{3D}}$ and $AP_{\text{BEV}}$ by 1.05 and 0.59, better than 0.38 and 0.18 increase induced by CCM module. These results demonstrate the validity of MCA, which selects more useful infrastructure features at different scales based on vehicle features with a cross-attention mechanism. FC is designed to eliminate redundant information included in features, while it can also improve detection performance. This is because FC module increases the depth of the whole network and introduces extra computation, which can be regarded as feature refinement.

%  We visualize the infrastructure image feature at different stages of the model to show the procedure of feature enhancement. We select the same feature channel of different features for visualization, including the largest scale received feature $f_{inf}^{0}$ from FC's output, the output of MCA module $f_{inf}$, and the output one of CCM module  $f_{inf}^{\prime}$. The results show that MCA and CCM can learn a better image representation from multi-scale image features, which focuses on the area of objects to be detected.

% \begin{figure}[ht]
% 	\centering  
% 	\includegraphics[width=0.9\linewidth]{Figures/Feature_VIS.png} 
% 	\caption{Visualization of infrastructure feature at different stages.
%  }  
% 	\label{fig:feature_vis}   
% \end{figure}

\begin{figure}[h]
	\centering  
	\includegraphics[width=0.9\linewidth]{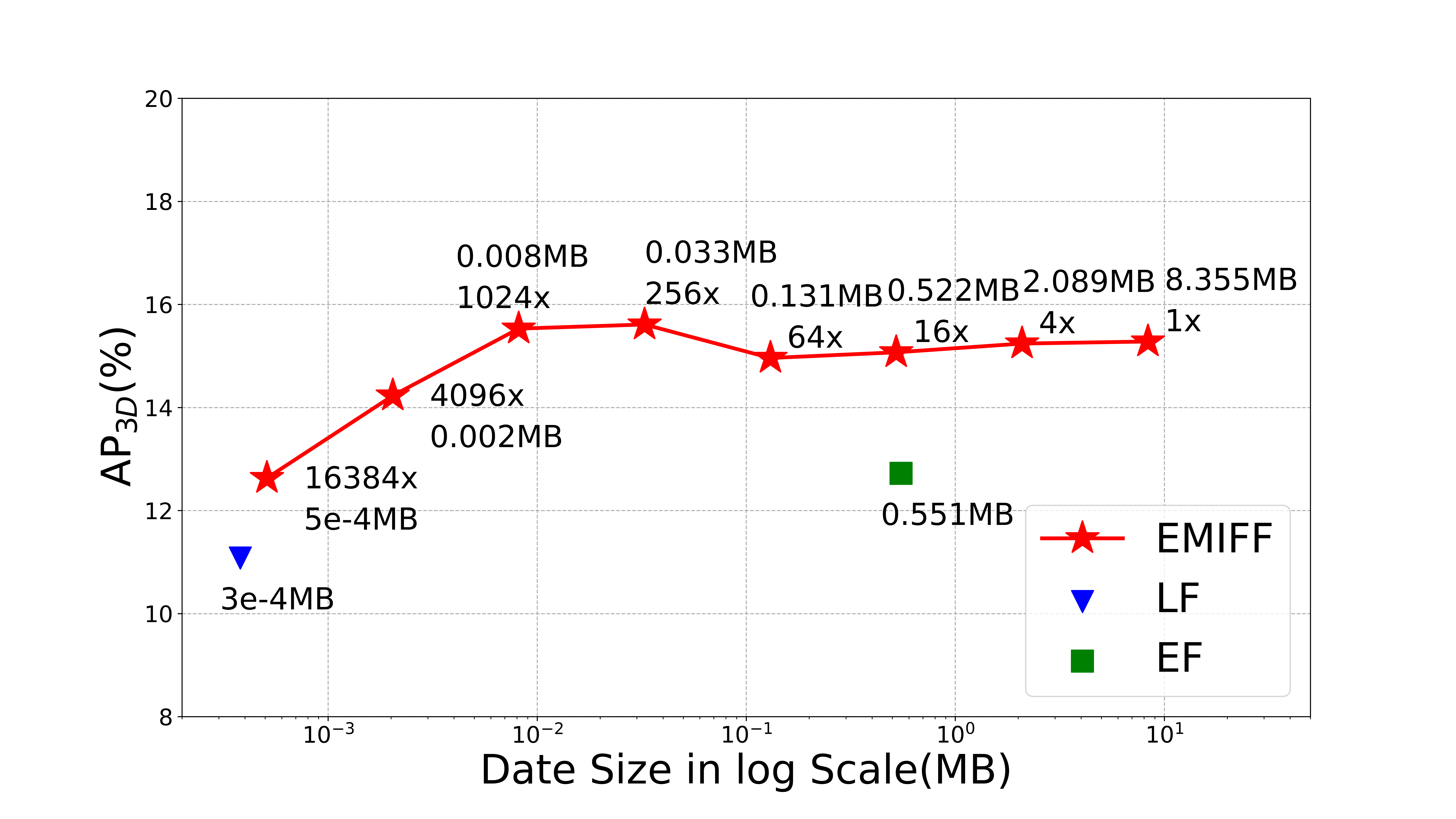} 
	\caption{$AP_{\text{3D (IoU=0.5)}}$ with respect to Compression Rate (shown as number $\times$). CCR is changed from $\times 1$ to $\times64$ and SCR is set from $\times 1$ to $\times 256$ with CCR set to $\times64$. 
 }  
	\label{fig:comp_3d}   
\end{figure}

\textbf{Voxel or BEV Fusion?} To investigate when to fuse features in IF method (at voxel or BEV level), we compare the performance of EMIFF with \textit{EMIFF\_BEV}. The former belongs to the IF-Voxel pipeline while the latter belongs to the IF-BEV fusion pipeline, which condenses voxel features $V_{veh}$ and $V_{inf}$ into BEV feature respectively with two 3D necks, and then two BEV features are averaged for fusion. Results (Table~\ref{TAB:VoxelBEV}) show that fusion at the voxel level has better performance, which indicates that the transformation from voxel to BEV feature can cause higher information loss.

\begin{table}[ht]
  \footnotesize
  \centering
    \begin{tabular}{cccc}
    \hline
    \textbf{Fusion} & \textbf{Model} & \bm{$AP_{\textbf{3D}}$} & \bm{$AP_{\textbf{BEV}}$}  \\ \hline
            LF       & ImVoxelNet & 11.08  & 14.76   \\ 
            EF       & ImVoxelNet\_M & \underline{12.72}  & \underline{18.17}   \\ 
            IF (BEV)  & EMIFF\_BEV & 11.50  & 16.23   \\ 
            IF (Voxel) & EMIFF & \textbf{13.37}  & \textbf{19.66}   \\ \hline
    \end{tabular}
    \caption{Analysis on choice of feature fusion.}
    \label{TAB:VoxelBEV}
\end{table}

\begin{table}[ht]
    % \footnotesize
    \centering
    \resizebox{\linewidth}{!}{
    \begin{tabular}{cccccc}
    \hline
    Backbone & Model & $AP_{3D}$ & $AP_{BEV}$ & Param & FLOPs \\ \hline
    ResNet-50 & EMIFF\_B & 13.60 & 20.05 & 47.82M & 94.01G \\
    ResNet-50 & EMIFF & 15.61 & 21.44 & 49.32M & 123.76G \\
    ResNet-101 & EMIFF\_B & 14.06 & 20.56 & 85.79M & 171.10G \\
    ResNet-101 & EMIFF & 16.46 & 22.32 & 87.31M & 201.46G \\ \hline
    \end{tabular}}
    \caption{Comparison results of model capacities.}
    \label{TAB:VIMI_FLOPs}
\end{table}

\subsection{Influence of Model Capacities}

To further explain the effect of the proposed modules, we replace the image backbone of EMIFF and the baseline model \textit{EMIFF\_B} from ResNet-50 to ResNet-101 to assess the extent of performance improvement that can be obtained by increasing the parameters and capacities. Results in Table~\ref{TAB:VIMI_FLOPs} show that a more complex network with higher capacity has better performance, while EMIFF with fewer parameters and lower FLOPs still outperforms baseline \textit{EMIFF\_B} with ResNet-101. This observation provides additional evidence supporting the effectiveness of the proposed modules.

\subsection{Impact of Feature Compression}

As seen in Figure~\ref{fig:comp_3d}, We investigate the effect of Channel Compressor and Spatial Compressor. First, we change Channel Compression Rate (CCR) from $\times1$ to $\times64$, and the model performance is almost stable at low compression rates, which indicates that channel compression can extract more useful information and remove redundancy. After CCR reaches the maximum, we continue to compress features with Spatial Compressor. The compression rate ranges from $\times64$ to $\times 16384$. With compressed feature shapes getting smaller, the $AP_{\text{3D}}$ declines from 15.33 to 12.63 but is still higher than LF, and the transmission cost has fallen to 0.51KB which is comparable to LF's cost.

\section{CONCLUSIONS}

EMIFF is a novel multi-view intermediate-fusion framework for camera-based VIC3D task. To correct the negative effect of pose errors and time asynchrony, we design a Multi-scale Cross-Attention module and Camera-aware Channel Masking module to fuse and augment multi-view features. EMIFF also effectively reduces transmission cost via Feature Compression, and has achieved state-of-the-art results on DAIR-V2X-C benchmark, significantly outperforming previous EF and LF methods. Future study points to extension of the framework to more data modalities.

% \addtolength{\textheight}{-12cm}   % This command serves to balance the column lengths
%                                   % on the last page of the document manually. It shortens
%                                   % the textheight of the last page by a suitable amount.
%                                   % This command does not take effect until the next page
%                                   % so it should come on the page before the last. Make
%                                   % sure that you do not shorten the textheight too much.

%%%%%%%%%%%%%%%%%%%%%%%%%%%%%%%%%%%%%%%%%%%%%%%%%%%%%%%%%%%%%%%%%%%%%%%%%%%%%%%%

%%%%%%%%%%%%%%%%%%%%%%%%%%%%%%%%%%%%%%%%%%%%%%%%%%%%%%%%%%%%%%%%%%%%%%%%%%%%%%%%

%%%%%%%%%%%%%%%%%%%%%%%%%%%%%%%%%%%%%%%%%%%%%%%%%%%%%%%%%%%%%%%%%%%%%%%%%%%%%%%%

\section*{ACKNOWLEDGMENT}
This work is funded by the National Key R\&D Program of China (2022ZD0115502) and Lenovo Research.

% %%%%%%%%%%%%%%%%%%%%%%%%%%%%%%%%%%%%%%%%%%%%%%%%%%%%%%%%%%%%%%%%%%%%%%%%%%%%%%%%

% References are important to the reader; therefore, each citation must be complete and correct. If at all possible, references should be commonly available publications.

\bibliographystyle{IEEEtran}
\balance
\bibliography{IEEEexample}

\end{document}